\documentclass{article}
\usepackage{amsmath}

\usepackage[numbers, super]{natbib}
\usepackage{threeparttable}

\makeatletter
\renewcommand\@biblabel[1]{#1.}
\makeatother
\usepackage{float}
\usepackage{graphicx}
\usepackage{enumitem} 
\usepackage[utf8]{inputenc} 
\usepackage[T1]{fontenc}    
\usepackage{hyperref}       
\usepackage{url}            
\usepackage{booktabs}       
\usepackage{amsfonts}       
\usepackage{nicefrac}       
\usepackage{microtype}      
\usepackage{xcolor}         

\usepackage[final]{neurips_2023}

\title{DynaGRAG | Exploring the Topology of Information for Advancing Language Understanding and Generation in Graph Retrieval-Augmented Generation}

\author{%
  Karishma Thakrar \\
  \texttt{karishma.thakrar@gatech.edu}
}

\begin{document}

\maketitle


\renewcommand{\thefootnote}{\fnsymbol{footnote}}
\footnotetext[1]{This architecture was first created in June 2024 and the paper was later published on arXiv in January 2025.}
\renewcommand{\thefootnote}{\arabic{footnote}}

\begin{abstract}

Graph Retrieval-Augmented Generation (GRAG or Graph RAG) architectures aim to enhance language understanding and generation by leveraging external knowledge. However, effectively capturing and integrating the rich semantic information present in textual and structured data remains a challenge. To address this, a novel GRAG framework, Dynamic Graph Retrieval-Agumented Generation (DynaGRAG), is proposed to focus on enhancing subgraph representation and diversity within the knowledge graph. By improving graph density, capturing entity and relation information more effectively, and dynamically prioritizing relevant and diverse subgraphs and information within them, the proposed approach enables a more comprehensive understanding of the underlying semantic structure. This is achieved through a combination of de-duplication processes, two-step mean pooling of embeddings, query-aware retrieval considering unique nodes, and a Dynamic Similarity-Aware BFS (DSA-BFS) traversal algorithm. Integrating Graph Convolutional Networks (GCNs) and Large Language Models (LLMs) through hard prompting further enhances the learning of rich node and edge representations while preserving the hierarchical subgraph structure. Experimental results demonstrate the effectiveness of DynaGRAG, showcasing the significance of enhanced subgraph representation and diversity for improved language understanding and generation.

\end{abstract}



\section{Introduction}

Recent breakthroughs in large language models have transformed natural language processing, demonstrating unparalleled capabilities in understanding and generating human-like text across diverse domains. Models like OpenAI o1, Claude 3 Opus, and Llama 70b have set new performance benchmarks, achieving state-of-the-art results.~\cite{vellum2025} However, tasks like analogical thinking, causal inference, and synthesizing complex knowledge continue to expose the limitations of these models, particularly in handling nuanced reasoning. To address these challenges, initiatives like OpenAI’s Project Strawberry~\cite{reuters2024} aim to empower LLMs with "human-like reasoning" by enabling them to autonomously explore, research, and plan. Even so, LLMs are fundamentally constrained by their reliance on static datasets, limiting their ability to dynamically adapt to evolving information. Fine-tuning, a common method for domain-specific adaptation, incurs significant computational costs and risks compromising reasoning capabilities.~\cite{Lobo:2024} This underscores the need for alternative approaches that deliver contextually rich and relevant responses while effectively leveraging and synthesizing complex information.

An emerging solution lies in the use of graph-structured data, particularly knowledge graphs, which capture entities and their relationships, offering a structured and enriched representation of information. These graphs have shown immense promise for improving domain-specific relevance and reasoning in tasks like question answering, recommendation systems, and natural language generation. However, unlocking the full potential of knowledge graphs requires overcoming a significant challenge: generating responses that effectively integrate the textual richness of LLMs with the topological insights of graph data. Achieving this synthesis necessitates rethinking how structured data is combined with generative AI models to produce scalable, relevant, and reasoning-rich outputs.


\begin{figure}[H]
    \centering
    \begin{minipage}[t]{0.45\linewidth}
        \centering
        \includegraphics[width=\linewidth]{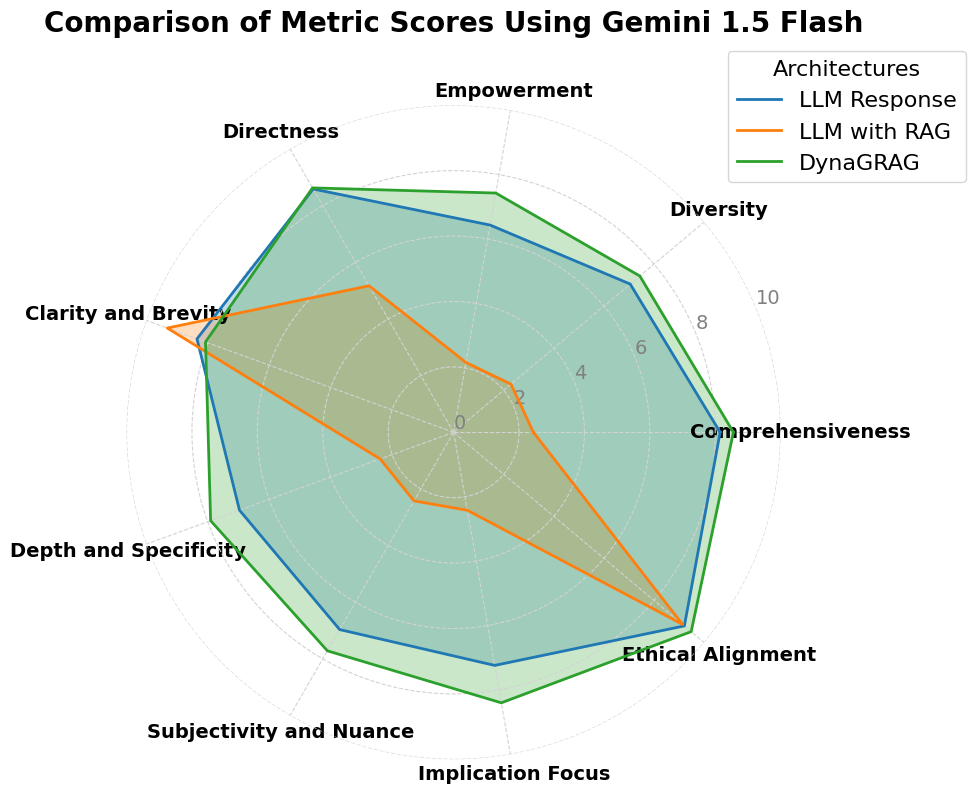}
        \caption{Radar plot illustrating performance across evaluation metrics for architectures tested using Gemini 1.5 Flash, highlighting DynaGRAG's strengths with empowerment, subjectivity and nuance, and implication focus.}
        \label{fig:radar_plot_gemini}
    \end{minipage}%
    \hfill
    \begin{minipage}[t]{0.5\linewidth}
        \centering
        \includegraphics[width=\linewidth]{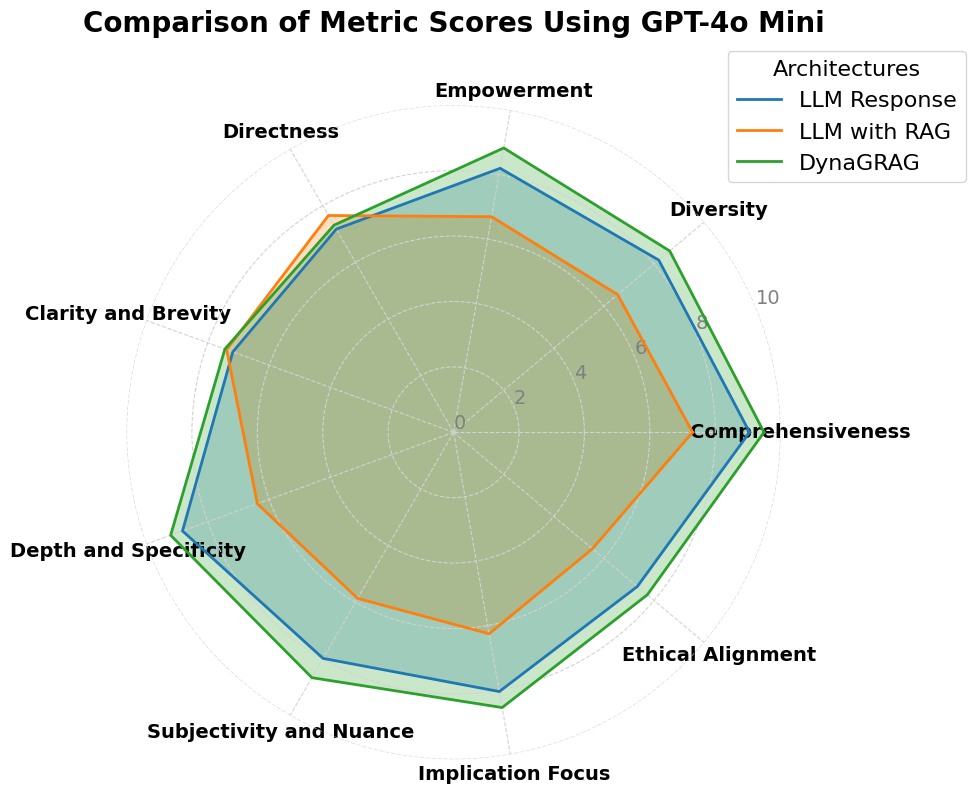}
        \caption{Radar plot illustrating performance across evaluation metrics for architectures tested using GPT 4-o mini, highlighting DynaGRAG's strengths across all metrics collectively.}
        \label{fig:radar_plot_openai}
    \end{minipage}
\end{figure}

One of the most notable advancements in integrating graph-structured data with generative AI models is Microsoft’s Graph RAG architecture proposed by Larson et al.~\cite{Edge:2024} This approach leveraged knowledge graphs to augment LLMs by generating pre-summarized community reports for distinct clusters of entities and relationships, enabling scalable query-focused summarization. While innovative in addressing large-scale global summarization, there are several limitations. The reliance on pre-summarized community reports sacrifices the granularity and flexibility of the underlying graph, reducing its ability to dynamically explore relationships or address queries requiring nuanced, unexpected connections. Moreover, the query-focused summarization approach consumes a substantial number of tokens, leading to significant computational overhead and higher costs during inference. If the structure of the underlying graph data changes, even slightly, the entire graph index and its community summaries would need to be regenerated, making the system inefficient and less adaptable to evolving datasets, a necessary feature in most practical applications of these graph-based algorithms.

Other frameworks, such as Graph RAG proposed by Hu et al.,~\cite{Hu:2024} a group of Emory University researchers, seek to improve retrieval-augmented generation by incorporating soft prompts and leveraging graph and text views. While conceptually appealing, this approach faces challenges with mainstream LLM architectures, which are not natively designed to process graph-based contextual information, limiting its practicality. Novel approaches that can capture and integrate the rich semantic information present in textual and structured data are needed to enable LLMs to reason over and generate responses from knowledge graphs, ultimately improving the depth and contextual relevance of AI-generated outputs.

The proposed Graph RAG architecture, DynaGRAG, fundamentally addresses the limitations of earlier approaches by preserving graph data in its native form, enabling real-time traversal and retrieval of entities, relationships, and their embeddings during query processing. Unlike static systems, this architecture dynamically assembles responses by exploring and integrating information directly from the graph. This ensures that the full richness and granularity of the data are preserved, allowing for flexible adaptation to evolving datasets and complex queries.

The architecture introduces several innovative features that enhance its effectiveness. It employs a process to improve graph density by consolidating redundant information and emphasizing the most meaningful nodes and edges, ensuring a clearer and more impactful graph representation. A unique two-step mean-pooling method captures diverse and comprehensive summaries of graph entities, enabling richer contextual understanding. Its query-aware retrieval mechanism retrieves subgraphs that are both highly relevant and diverse, achieving a balance between precision and variety. A dynamic traversal algorithm further refines the process by intelligently adapting the exploration of the graph, uncovering deeper and previously overlooked connections. By integrating graph neural networks to refine node and edge representations and leveraging the generative natural language capabilities of large language models, the architecture seamlessly bridges structured and unstructured data. Finally, carefully designed prompts ensure that both the textual richness and hierarchical insights of the graph are fully utilized, delivering exceptional performance on complex tasks requiring nuanced insights and multistep reasoning.

Through extensive experimentation, DynaGRAG has demonstrated significant improvements over existing methods. By combining the retrieval of graph-structured data with LLMs, it effectively captures complex relationships among seemingly unrelated information, traverses long contexts, and synthesizes nuanced, contextually rich responses. This integration of textual and topological insights not only enhances the system’s reasoning and adaptability but also has demonstrated aligned outputs without explicit directives for ethical considerations. By focusing on retrieving and analyzing relevant subgraphs and grounding responses in relationships and context, the framework consistently delivers results that are coherent, meaningful, and accountable. These findings underscore the transformative potential of robust reasoning techniques in fostering thoughtful and responsible AI systems.

\begin{figure}[H]
    \centering
    \includegraphics[width=0.85\linewidth]{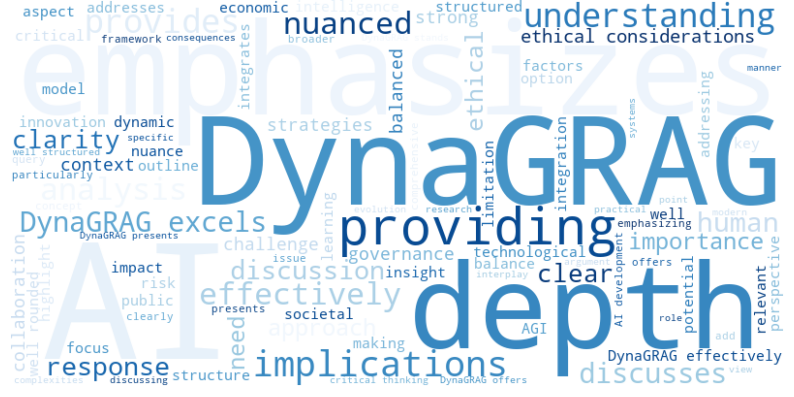}
    \caption{Word cloud representing an LLM's evaluation of the proposed framework's responses, highlighting attributes such as understanding, depth, clarity, focus on ethical and societal implications, and complex reasoning.}
    \label{fig:sample}
\end{figure}


\section{Related Work}


The integration of external knowledge with LLMs through RAG has gained significant recent attention. Gao et al. (2024)~\cite{Gao:2024} provide a comprehensive survey of RAG methods, discussing the evolution of general paradigms and core components. Graph-based approaches have been explored to augment LLMs in various domains, such as knowledge graph generation, (Chang et al., 2024; Yao et al., 2024)~\cite{Chang:2024}~\cite{Yao:2024} bidirectional reasoning (Pan et al., 2024)~\cite{Pan:2024} and categorization of graph generation (Guo and Zhao, 2022);~\cite{Guo:2022} these works primarily focus on broader integration strategies, specific graph generation tasks, or providing detailed taxonomies and evaluations of existing models. Several recently proposed GRAG frameworks target multi-hop reasoning on textual graphs (Hu et al., 2024),~\cite{Hu:2024} knowledge graph question answering (Mavromatis and Karypis, 2024),~\cite{Mavromatis:2024} relational knowledge (Peng et al., 2024),~\cite{Peng:2024} and addressing challenges such as information hallucination and catastrophic forgetting (Sanmartín et al., 2024).~\cite{Sanmartin:2024} While these approaches address specific challenges, they often face compatibility issues with mainstream LLMs, prioritize dense subgraph reasoning and factual accuracy over capturing a richer variety of entity relationships, or focus on formalizing and surveying existing methods. Larson et al. (2024)~\cite{Edge:2024} propose a Graph RAG approach for query-focused summarization that relies on multi-level community summaries. While thorough, the overall framework requires frequent regeneration of summaries as the underlying data evolves. Other works explore encoding graph-structured data as text for LLMs to improve graph reasoning tasks (Fatemi et al., 2023)~\cite{Fatemi:2023} or investigate LLMs for learning on text-attributed graphs (Chen et al., 2024).~\cite{Chen:2024} These papers primarily address encoding strategies, graph reasoning challenges in black-box LLM environments, or emphasize node classification tasks, differing from the focus on enhancing subgraph diversity and representation for improved language understanding and generation.


To overcome the limitations identified in existing approaches, this work introduces a novel GRAG framework designed to enhance subgraph diversity and representation within knowledge graphs. A few key innovations include techniques such as a two-step de-duplication for consolidating similar entities and relationships, query-aware retrieval that prioritizes unique nodes, and a Dynamic Similarity-Aware BFS traversal algorithm to uncover deeper contextual connections. These advancements enable more nuanced and comprehensive language understanding and generation, extending the capabilities of existing RAG and GRAG methodologies.

\section{Methodology}

The first step in DynaGRAG is to construct a knowledge graph from textual data. This process involves dividing source documents into text chunks and identifying entities and their relations within these chunks. An LLM is used to extract significant entities, their summaries, and the relationships between them. The extracted information is then used to generate embeddings for the entities and relations using a pre-trained language model (PLM). This approach to knowledge graph construction enables user-specified LLM prompts tailored to specific domains or themes, allowing for the flexible extraction and prioritization of relevant information from source documents.



\begin{figure}[H]
    \centering
    \includegraphics[width=0.85\linewidth]{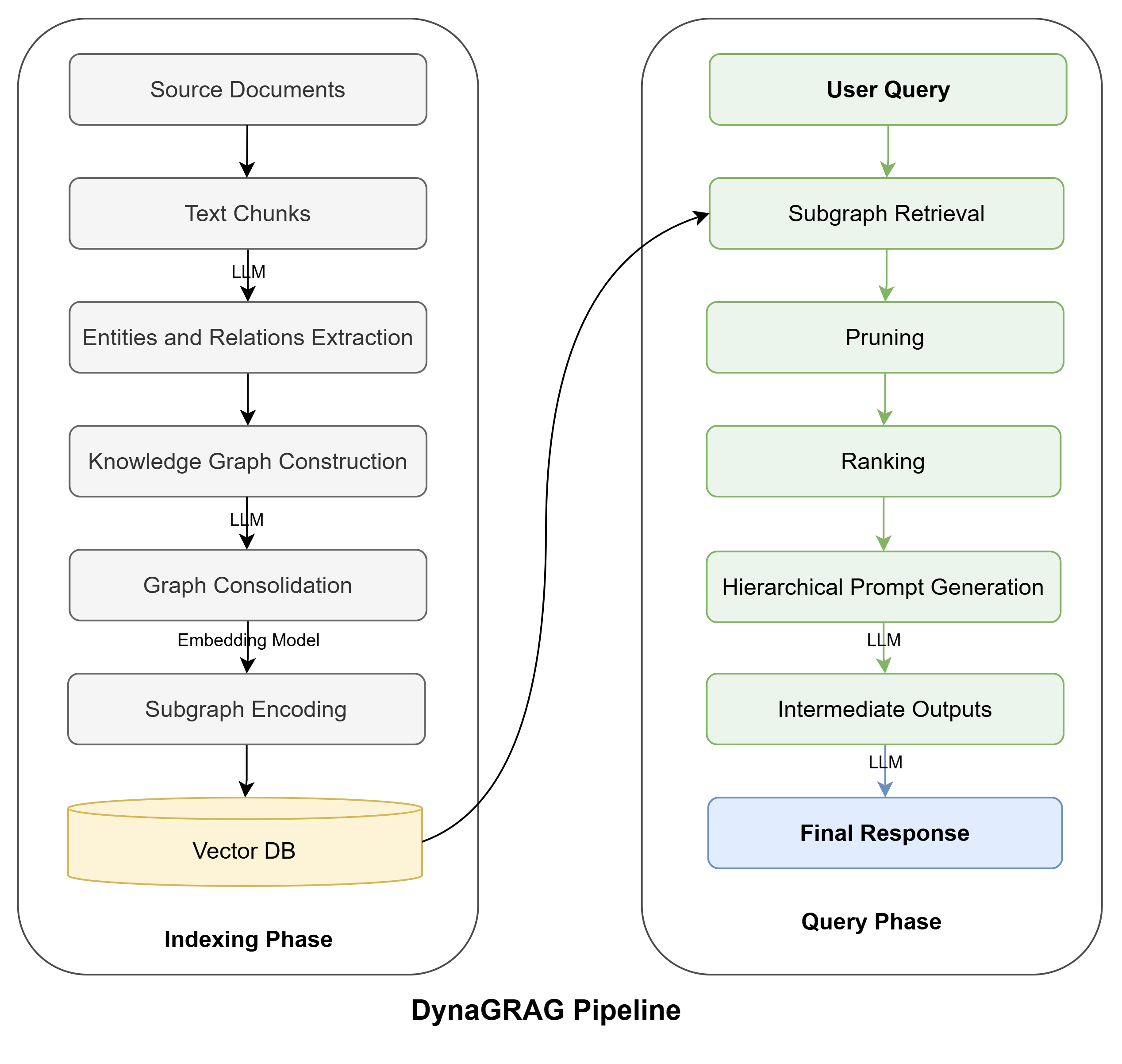}
    \caption{Process diagram describing the graph retrieval and generation process in DynaGRAG.}
    \label{fig:sample}
\end{figure}

\subsection{Graph Consolidation}
To enhance the graph representation of data, several novel techniques are employed to improve graph connectivity and capture the prevalence of entities and relations. The LLM identifies highly similar or synonymous entities by analyzing the semantic similarity between entity labels and their context within the graph, aligning the graph with human understanding. For example, entities such as "compute", "compute resources", and "compute resources usage" can be consolidated into a single entity, like "compute". This consolidation process effectively handles new terms like "GPT-7", which models like Word2Vec often lack vector representations for.


To further enhance the representation of information, the de-duplication process involves a two-step approach to averaging entity embeddings. First, the embeddings of identical entities are averaged, ensuring that each unique entity is represented accurately. Then, the embeddings of similar entities averaged, capturing the variety of entity representations without overemphasizing the prevalence of a single version. This is meaningful as often times LLMs produce identical summaries for the same entities in a chunk and so following this two-step averaging process ensures the capturing of a variety of entity summary representations more effectively. A key advantage of this approach is its ability to balance representation by preventing overemphasis on high-frequency entities, ensuring that less common but meaningful entities are not overshadowed, and the final output reflects an unbiased average of distinct latent features. 

The de-duplication process results in condensed entities and relations, which form the basis of undirected weighted graphs. By explicitly handling duplicates, critical information such as the relative importance of nodes and edges are captured, storing their frequency of occurrence as weights and accumulating all relation and entity summaries. This approach enhances graph density, resulting in a more interconnected and less sparse graph structure that allows for efficient propagation of information, ultimately improving retrieval performance in the GRAG architecture. Moreover, the de-duplication process is flexible and can be controlled in terms of the granularity sought, allowing users to tailor and optimize the graph consolidation process for the unique characteristics of each dataset and application and enhancing the graph representation for various downstream tasks.

\begin{figure}
	\centering
	\includegraphics[width=0.8\linewidth]{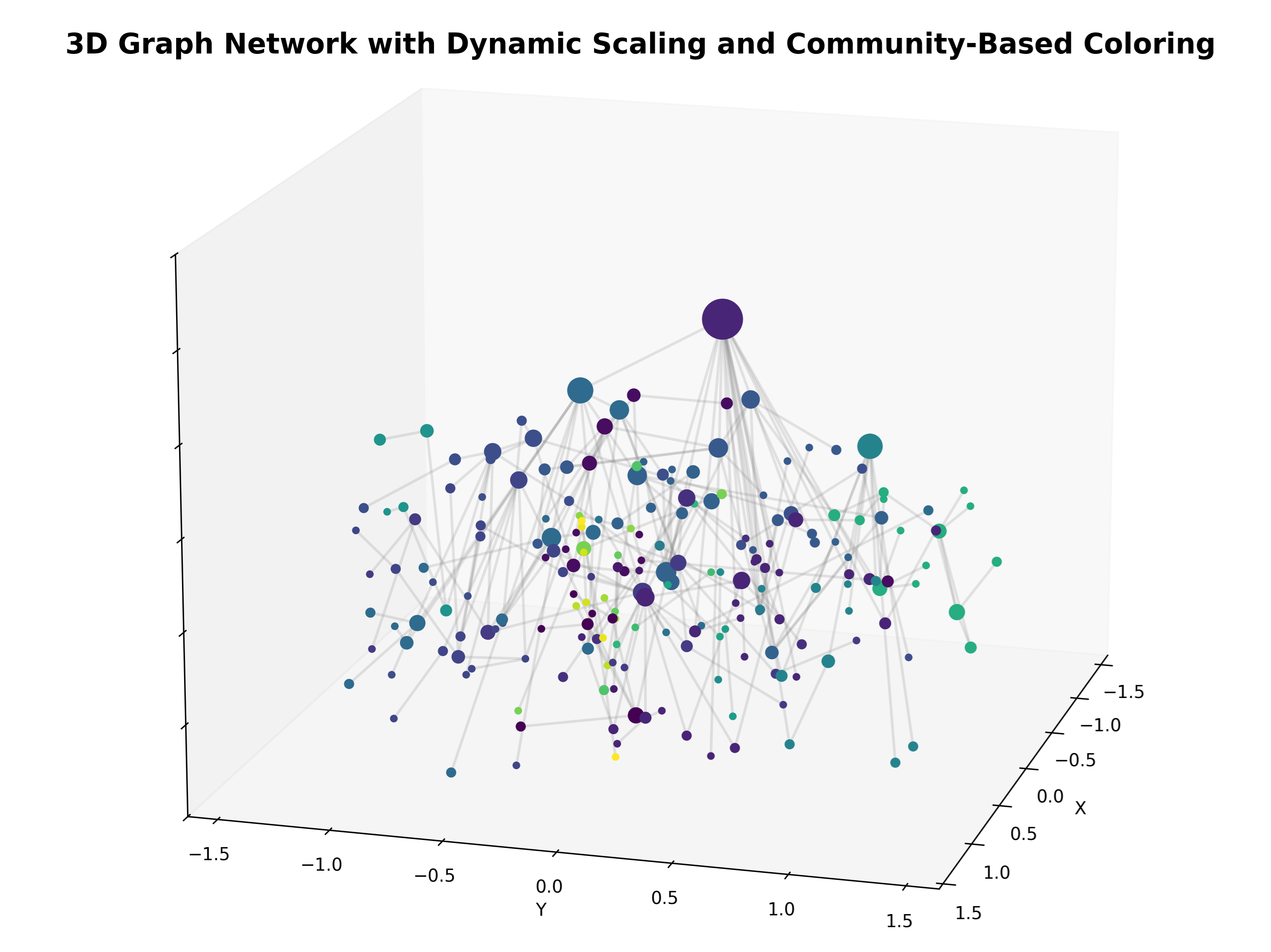}
    \caption{Visualization of a knowledge graph extended to a 3D representation based on nodes and entities extracted from transcripts of Dwarkesh Patel's podcast.\cite{DwarkeshPodcast:2025} Nodes are positioned using the Kamada-Kawai layout, dynamically scaled, and perturbed along the z-axis for enhanced visual differentiation.}
	\label{fig:sample}
\end{figure}

\subsection{Subgraph Retrieval}
To efficiently retrieve relevant information from a large graph structure, the system constructs a searchable database of pre-computed ego-graphs centered on each node, including nodes within k hops (typically, k=3). These ego-graphs are encoded by first computing weighted node and edge features to capture the structural and semantic significance of each entity and relationship.

\vspace{1em}

\textbf{Weighted Node Features:}
\[
\mathbf{h}_i^{\text{weighted}} = \mathbf{h}_i \cdot w_i
\]
Where:
\begin{itemize}
    \item \(\mathbf{h}_i\): Node embedding for node \(i\)
    \item \(w_i\): Weight (i.e., entity weight) for node \(i\)
    \item \(\mathbf{h}_i^{\text{weighted}}\): Weighted node feature
\end{itemize}

\vspace{1em}

\textbf{Weighted Edge Features:}
\[
\mathbf{e}_{ij}^{\text{weighted}} = \frac{\mathbf{h}_i + \mathbf{h}_k + \mathbf{r}_{ij}}{n} \cdot w_{ij}
\]
Where: 
\begin{itemize}
    \item \( \mathbf{h}_i \): Source embedding (node \(i\)).
    \item \( \mathbf{h}_k \): Target embedding (node \(k\)).
    \item \( \mathbf{r}_{ij} \): Relationship embedding for the edge between nodes \(i\) and \(k\).
    \item \( n \): Number of embeddings being averaged (i.e., \(n = 3\) for source, target, and relationship).
    \item \( w_{ij} \): Relationship weight for the edge.
\end{itemize}

Once the weighted node and edge features are computed, they are combined to form the total graph embedding, which represents the encoded ego-graph as:

\vspace{1em}

\textbf{Total Graph Embedding:}
\[
\mathbf{g} = \frac{\sum_{i} \mathbf{h}_i^{\text{weighted}} + \sum_{(i,j)} \mathbf{e}_{ij}^{\text{weighted}}}{w_{\text{total}}}
\]

Where: 
\begin{itemize}
    \item \( \mathbf{g} \): Graph embedding.
    \item \( \mathbf{h}_i^{\text{weighted}} \): Weighted node feature for node \(i\), derived as \( \mathbf{h}_i \cdot w_i \) (node embedding multiplied by its weight).
    \item \( \mathbf{e}_{ij}^{\text{weighted}} \): Weighted edge feature for edge \((i, j)\), as derived earlier.
    \item \( w_{\text{total}} \): Total weight, which is the sum of all node weights (\( \sum_i w_i \)) and edge weights (\( \sum_{(i,j)} w_{ij} \)).
\end{itemize}

\vspace{1em}


During retrieval, the total graph embedding of each pre-computed ego-graph is compared to the query embedding using cosine similarity. The top-N subgraphs are ranked based on their similarity scores, ensuring relevance to the query. To enhance diversity, a mechanism tracks the top nodes of previously retrieved subgraphs and iteratively adds subgraphs with distinct key entities. This strategy avoids redundancy and provides a richer, more balanced set of results by selecting subgraphs that span different regions of the knowledge graph.

To assess the effectiveness of diversity-aware retrieval, its impact on intermediate results for specific queries was analyzed. Without the diversity criterion, the system frequently prioritized a narrow set of dominant subgraphs focused on highly interconnected entities. For example, queries about the "Future of AI" consistently retrieved subgraphs exploring similar themes like such as "AI researchers" or "Bias in artificial intelligence."  While relevant, this approach had disregarded less prominent but also meaningful subgraphs. The introduction of a diversity criterion fundamentally transformed the retrieval process, surfacing a wider array of subgraphs for this query, such as Geoffrey Hinton’s concerns about AI, RLHF’s implications for safety, and advances in attention mechanisms. By penalizing subgraphs with overlapping top nodes, the system captured distinct regions of the knowledge graph, uncovering previously underexplored insights and enhancing the system's ability to address complex, multifaceted questions with nuanced and comprehensive responses.

\begin{figure}[H]
    \centering
    \includegraphics[width=0.85\linewidth]{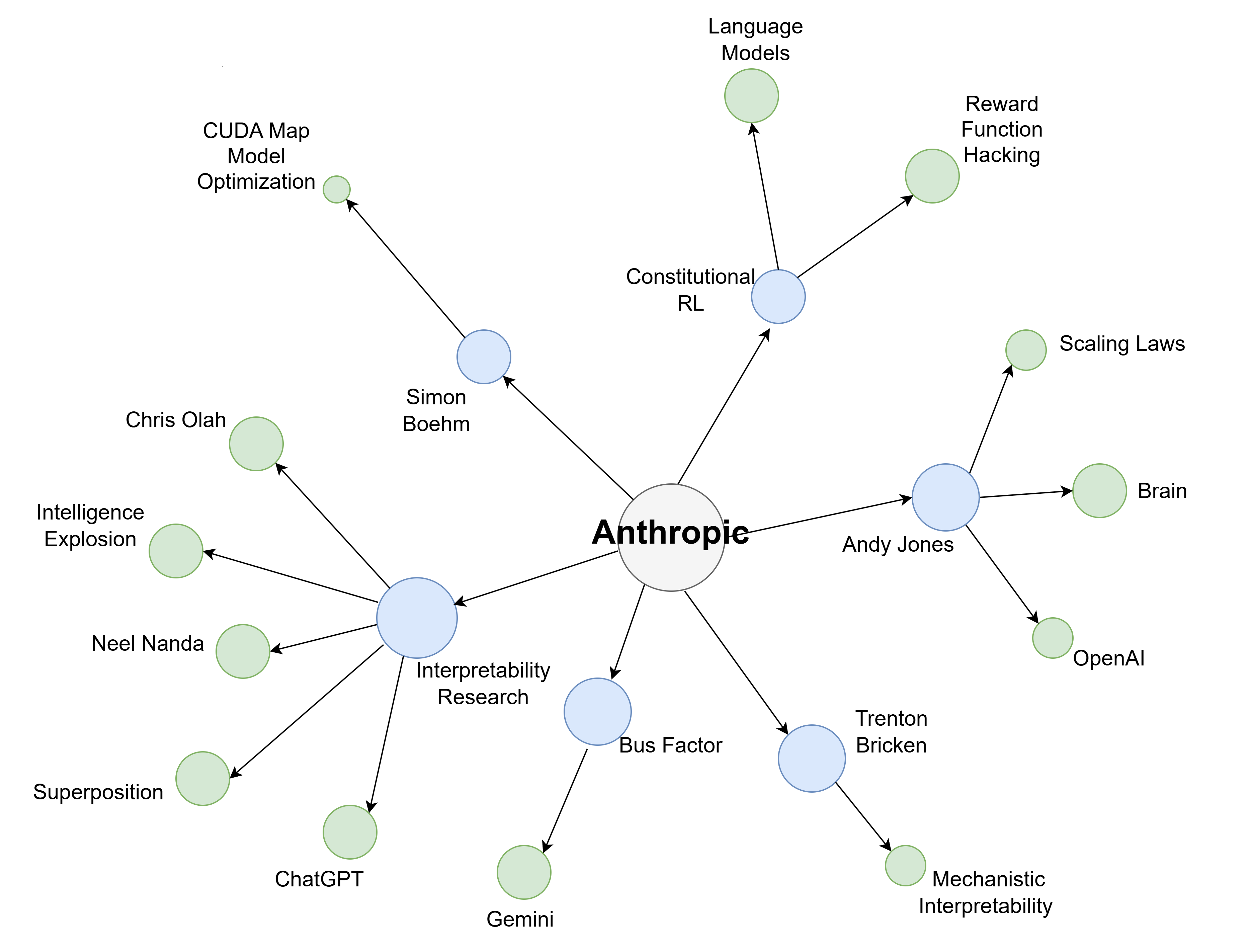}
    \caption{Example 2-hop ego-network for Anthropic from Dwarkesh Patel's Podcast episode with Sholto Douglas \& Trenton Bricken.\cite{DwarkeshPodcast:2025} Node sizes reflecting their weights and prevalence within the subgraph.}
    \label{fig:sample}
\end{figure}

\subsection{Subgraph Pruning}

The retrieved subgraphs are refined to prioritize components most relevant to the query. This process begins with the computation of prominence scores, which combine structural significance and semantic alignment. Structural significance is determined using pre-computed weights for entities and relationships, reflecting factors such as centrality or connectivity. Semantic alignment, on the other hand, is calculated as the Euclidean distance between the query embedding and the embeddings of nodes or edges. These distances are transformed into normalized relevance scores using a multilayer perceptron (MLP) with a sigmoid activation function, assigning higher scores to elements that are more closely aligned with the query while ensuring consistent scaling for downstream use.



\vspace{1em}

\textbf{Node-Query Distance (Euclidean Distance):}
\[
\text{Distance}_{\text{node}} = \|\mathbf{h}_i - \mathbf{q}\|_2 = \sqrt{\sum_{j=1}^d (h_{i,j} - q_j)^2}
\]

Where:
\begin{itemize}
    \item \( \mathbf{h}_i \): Node embedding for node \(i\) (a \(d\)-dimensional vector).
    \item \( \mathbf{q} \): Query embedding (a \(d\)-dimensional vector).
    \item \( d \): Dimensionality of the embedding space.
    \item \( h_{i,j} \): \(j\)-th component of the node embedding.
    \item \( q_j \): \(j\)-th component of the query embedding.
\end{itemize}

\vspace{1em}

\textbf{Edge-Query Distance (Euclidean Distance):}
\[
\text{Distance}_{\text{edge}} = \|\mathbf{r}_{ij} - \mathbf{q}\|_2 = \sqrt{\sum_{j=1}^d (r_{ij,j} - q_j)^2}
\]

\begin{itemize}
    \item \( \mathbf{r}_{ij} \): Relationship embedding (edge embedding) for the edge between nodes \(i\) and \(j\) (a \(d\)-dimensional vector).
    \item \( \mathbf{q} \): Query embedding (a \(d\)-dimensional vector).
    \item \( d \): Dimensionality of the embedding space.
    \item \( r_{ij,j} \): \(j\)-th component of the edge embedding.
    \item \( q_j \): \(j\)-th component of the query embedding.
\end{itemize}

Once relevance scores are computed, they are passed into a Graph Convolutional Network (GCN) to generate updated pruning scores. The GCN incorporates both the initial relevance scores and the structural relationships within the graph, dynamically adjusting the importance of nodes and edges. Rather than applying hard masking to remove less relevant components, the framework employs soft masking to scale their influence. Node weights are adjusted based on their pruning scores, and edge attributes are similarly scaled to reflect their updated relevance. Through its message-passing process, the GCN also further refines the node and edge representations by aggregating information from their neighbors. The result of this process is a refined subgraph that is contextually rich and closely aligned with the query. These subgraphs are optimized for downstream tasks such as reasoning and generation, effectively leveraging both the structural complexity and semantic depth of the knowledge graph.

\subsection{Hard Prompting}

In the DynaGRAG framework, hard prompts are generated by traversing the pruned subgraphs. A novel Dynamic Similarity-Aware BFS algorithm is implemented, which adjusts the node exploration order based on similarity scores, prioritizing the exploration of highly similar neighbors. This innovative approach reveals deeper connections between nodes that might be missed in a strict BFS, enhancing the coherence of the prompt while maintaining the logical structure. Pre-order traversal algorithms further enrich the prompt by collecting relevant information, such as entity summaries, edge relationships, and their pruned weights, providing additional context about the importance of each entity and relationship within the subgraph. Non-tree edges are also processed, offering supplementary insights and connections beyond the basic tree structure. The resulting prompt string reflects the hierarchical structure of the subgraph, aligning with the strengths of LLMs, which are pre-trained on hierarchical data formats such as JSON, XML, and nested outlines. By generating informative and structured prompts that capture both the textual and topological aspects of the knowledge graph, DynaGRAG leverages the strengths of LLMs in understanding and reasoning about hierarchical data.


\subsection{Generating Responses}
The query and hard prompt are combined into a single input string, which is then used to generate intermediate responses from each retrieved subgraph. These intermediate responses capture different aspects and perspectives related to the query from various parts of the knowledge graph. The system calculates a helpfulness score for each intermediate response based on relevance, coherence, and level of detail. The intermediate responses are then sorted by their helpfulness scores in descending order. The sorted intermediate responses are fed into the language model, which generates a coherent and comprehensive final response, integrating information from multiple subgraphs while avoiding redundancy. This approach effectively encourages the LLM to "think" first about a question before forming its final response while ensuring that the generation process maintains the essential information from the input graph.

To summarize, this research proposes a novel GRAG framework that seamlessly integrates LLMs and graph learning techniques, introducing several key innovations:

\begin{enumerate}[leftmargin=0.5cm] 
    \item \textbf{Enhanced Graph Representation with De-duplication:} The framework employs a robust de-duplication process to consolidate similar entities and relationships, improving graph density and capturing the relative importance of nodes and edges. A two-step mean pooling of embeddings is applied to preserve diverse entity summary representations, ensuring a rich and balanced graph representation.
    \item \textbf{Query-aware Subgraph Retrieval with Diversity Prioritization:} A retrieval process ensures that subgraphs are not only relevant to the query but also diverse, by considering the uniqueness of top nodes within each subgraph. This balance prevents redundancy, surfaces underrepresented insights, and enriches the contextual breadth of retrieved subgraphs.
    \item \textbf{Dynamic Similarity-Aware BFS Algorithm:} A novel traversal method dynamically adjusts node exploration order based on similarity scores. This approach uncovers deeper, contextually meaningful connections that might be missed in a traditional BFS, while maintaining a balance between structural coherence and exploration flexibility.
    \item \textbf{Integration of Graph Convolutional Networks:} GCNs are utilized to learn rich representations for nodes and edges, leveraging both structural significance and semantic alignment to produce relevance scores. This process enhances the framework’s ability to refine subgraphs by emphasizing elements that are most critical to the query.
    \item \textbf{Harnessing LLMs for Hierarchical Data:} The framework capitalizes on LLMs' ability to reason about hierarchical data by generating hard prompts that preserve both textual and topological aspects of subgraphs. These structured prompts align with the strengths of LLMs, facilitating more coherent and contextually grounded responses.
\end{enumerate}






The framework employs several key strategies, including: (1) utilizing LLMs to enhance textual attributes of nodes, compensating for deficiencies in contextualized knowledge and semantic comprehension inherent in conventional GNN pipelines; and (2) adapting LLMs to generate informative, domain-aligned responses by directly leveraging the graph structure through techniques like hard prompting and intermediate answer generation.


\section{Experiments}

This study evaluated the performance of various LLM architectures in generating nuanced, contextually aware responses to queries based on the 2024 Dwarkesh Patel Podcast transcripts.\cite{DwarkeshPodcast:2025}

\subsection{Dataset and Preprocessing}
The dataset consisted of 460k tokens extracted from the 2024 Dwarkesh Patel Podcast transcripts. A total of 180 non-factoid queries were generated, balanced across episodes, to comprehensively test reasoning capabilities and cover a wide range of topics discussed in the podcasts.

\subsection{Architectures and Pipelines}
The evaluation utilized three distinct pipelines. Both GPT-4o mini and Gemini 1.5 Flash models were evaluated with each architecture.

\textbf{LLM Pipeline (Vanilla LLM):} In this pipeline, the queries were directly fed into the LLMs (GPT-4o Mini or Gemini 1.5 Flash) without any additional context or augmentation.
The primary goal was to evaluate the inherent reasoning, contextual interpretation, and response generation abilities of the models without external support.
The outputs reflected each model’s standalone performance, relying solely on their internal knowledge and inference capabilities.

\textbf{LLM with Retrieval-Augmented Generation Pipeline (Naïve RAG):} This pipeline combined LLMs with a robust retrieval system for contextual grounding. Podcast transcripts were segmented into 2,400-token chunks with a 200-token overlap for continuity. Chunks were embedded using JinaAI's jina-embeddings-v2-base-en model, supporting a context window of 8,192 tokens, and stored in a LanceDB vector store for efficient cosine similarity-based retrieval. For each query, the top five relevant chunks were retrieved and appended to the query, providing context to enhance the relevance of the LLM-generated responses.


\textbf{Dynamic Retrieval-Augmented Generation Pipeline (DynaGRAG):} DynaGRAG integrated graph-based retrieval, subgraph pruning, and advanced prompting to enhance response relevance and coherence. Podcast transcript chunks were converted into knowledge graph representations, with entities and relationships as nodes and edges. The pipeline used the Dynamic Similarity-Aware BFS algorithm to prioritize the exploration of similar nodes during subgraph retrieval, ensuring diversity and relevance. Key entities were highlighted, and irrelevant information was pruned separately using learned relevance scores. These refined subgraphs formed hierarchical prompts, aligning with the LLM's strengths to generate comprehensive, contextual responses.

\subsection{Evaluation Framework}
Each query response from the architecture-LLM pipeline was assessed using the following nine metrics designed to evaluate both the depth and quality of the responses:

\begin{itemize}[leftmargin=.5cm] 
    \item \textbf{Comprehensiveness:} Measures how effectively the response addresses all aspects of the query, emphasizing integration and coherence over isolated points.
    \item \textbf{Diversity:} Evaluates the breadth of perspectives, examples, or approaches provided, prioritizing responses that synthesize varied elements into a unified argument.
    \item \textbf{Empowerment:} Assesses the extent to which the response enhances understanding by offering actionable insights and fostering confidence in complex reasoning.
    \item \textbf{Directness:} Examines the response’s ability to stay focused on the query while exploring necessary complexities, striking a balance between clarity and depth as appropriate.
    \item \textbf{Clarity and Brevity:} Rates the ability to convey intricate reasoning in a concise and accessible manner, without oversimplifying or omitting critical details.
    \item \textbf{Depth and Specificity:} Evaluates the precision and relevance of reasoning, particularly how well the response uses detailed evidence or examples to substantiate its claims.
    \item \textbf{Subjectivity and Nuance:} Measures the ability to explore subtle distinctions, conflicting perspectives, and interdependencies, avoiding overly binary interpretations.
    \item \textbf{Implication Focus:} Assesses how thoroughly the response identifies and analyzes broader implications, including societal, ethical, or practical consequences.
    \item \textbf{Ethical Alignment:} Evaluates how well ethical considerations are integrated into the response, particularly their role in shaping or challenging the argument.
\end{itemize}

The first four metrics, Comprehensiveness, Diversity, Empowerment, and Directness, were inspired by Microsoft's GRAG evaluation framework. To extend this foundation, five additional metrics were introduced to provide a more holistic assessment of the LLMs’ reasoning, contextual understanding, and response quality. The scores for each pipeline were averaged and analyzed to identify trends and compare the relative strengths of the architecture-LLM combinations. 




\newpage

\section{Results}

The evaluation results revealed clear distinctions in the performance of the three architectures: LLM Response (Vanilla LLM), LLM with RAG (Naïve RAG), and DynaGRAG. Across both the Gemini and OpenAI models, DynaGRAG emerged as the most effective pipeline, excelling in metrics like Comprehensiveness (8.56 for Gemini, 9.49 for OpenAI), Diversity (7.43 for Gemini, 8.63 for OpenAI), and Depth and Specificity (7.91 for Gemini, 9.21 for OpenAI). These scores highlight DynaGRAG’s ability to capture nuanced, multifaceted information and provide diverse, contextually rich responses.

While the Vanilla LLM achieved good performance in Ethical Alignment (9.21 for Gemini, 7.34 for OpenAI), it struggled to match DynaGRAG in terms of reasoning depth and diversity, reflecting the limitations of relying solely on inherent model knowledge. The Naïve RAG pipeline, despite leveraging retrieval for context, exhibited lower performance across most metrics, with significant gaps in Diversity and Depth, indicating challenges in effectively incorporating retrieved information without graph representation into responses with reasoning and contextual understanding.


The overall average reasoning scores further underscore DynaGRAG’s effectiveness, achieving 8.18 for Gemini and 8.43 for OpenAI, compared to 7.66 and 7.99 for the Vanilla LLM, and 4.20 and 6.63 for the Naive RAG. These results emphasize the impact of DynaGRAG's graph-based retrieval and dynamic prompting techniques, which enable the architecture to integrate complex relationships, prioritize relevance, and deliver responses that align closely with the query’s context. The findings validate the strategic advantage of graph-structured approaches in advancing reasoning, diversity, and context integration in LLM pipelines.


\begin{table}[H]
\centering
\caption{Evaluation Metrics for Different Architectures}
\label{tab:evaluation_results_swapped}
\begin{tabular}{|l|c|c|c|}
\hline
\textbf{Metric} & \textbf{Vanilla LLM} & \textbf{Naïve RAG} & \textbf{DynaGRAG} \\ \hline
\multicolumn{4}{|c|}{\textbf{Gemini 1.5 Flash Results}} \\ \hline
Comprehensiveness & 8.15 & 2.44 & 8.56 \\ \hline
Diversity & 7.05 & 2.29 & 7.43 \\ \hline
Empowerment & 6.43 & 2.17 & 7.43 \\ \hline
Directness & 8.59 & 5.17 & 8.63 \\ \hline
Clarity and Brevity & 8.35 & 9.32 & 8.07 \\ \hline
Depth and Specificity & 6.97 & 2.39 & 7.91 \\ \hline
Subjectivity and Nuance & 6.97 & 2.42 & 7.72 \\ \hline
Implication Focus & 7.24 & 2.42 & 8.40 \\ \hline
Ethical Alignment & 9.21 & 9.14 & 9.48 \\ \hline
\textbf{Overall Reasoning Score} & \textbf{7.66} & \textbf{4.20} & \textbf{8.18} \\ \hline
\multicolumn{4}{|c|}{\textbf{GPT 4o-mini Results}} \\ \hline
Comprehensiveness & 9.05 & 7.31 & 9.49 \\ \hline
Diversity & 8.19 & 6.55 & 8.63 \\ \hline
Empowerment & 8.20 & 6.69 & 8.83 \\ \hline
Directness & 7.17 & 7.65 & 7.31 \\ \hline
Clarity and Brevity & 7.18 & 7.39 & 7.44 \\ \hline
Depth and Specificity & 8.82 & 6.39 & 9.21 \\ \hline
Subjectivity and Nuance & 7.98 & 5.87 & 8.66 \\ \hline
Implication Focus & 8.05 & 6.26 & 8.55 \\ \hline
Ethical Alignment & 7.34 & 5.54 & 7.73 \\ \hline
\textbf{Overall Reasoning Score} & \textbf{7.99} & \textbf{6.63} & \textbf{8.43} \\ \hline
\end{tabular}
\end{table}

\begin{figure}[H]
    \centering
    \begin{minipage}[t]{1\linewidth}
        \centering
        \includegraphics[width=\linewidth]{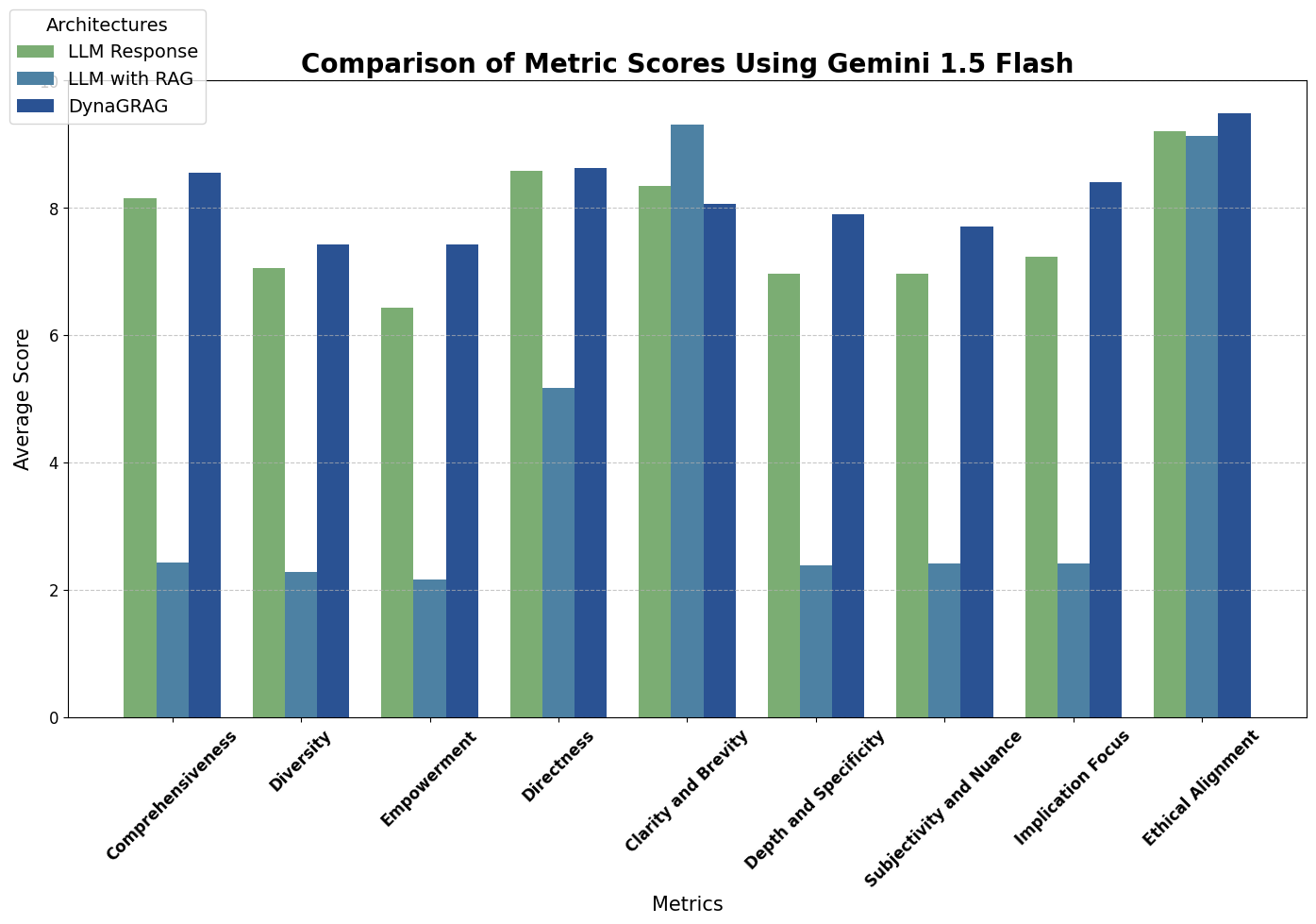}
        \caption{Bar plot of architecture results using Gemini 1.5 Flash.}
        \label{fig:gemini_results}
    \end{minipage}%
    \hfill
    \begin{minipage}[t]{1\linewidth}
        \centering
        \includegraphics[width=\linewidth]{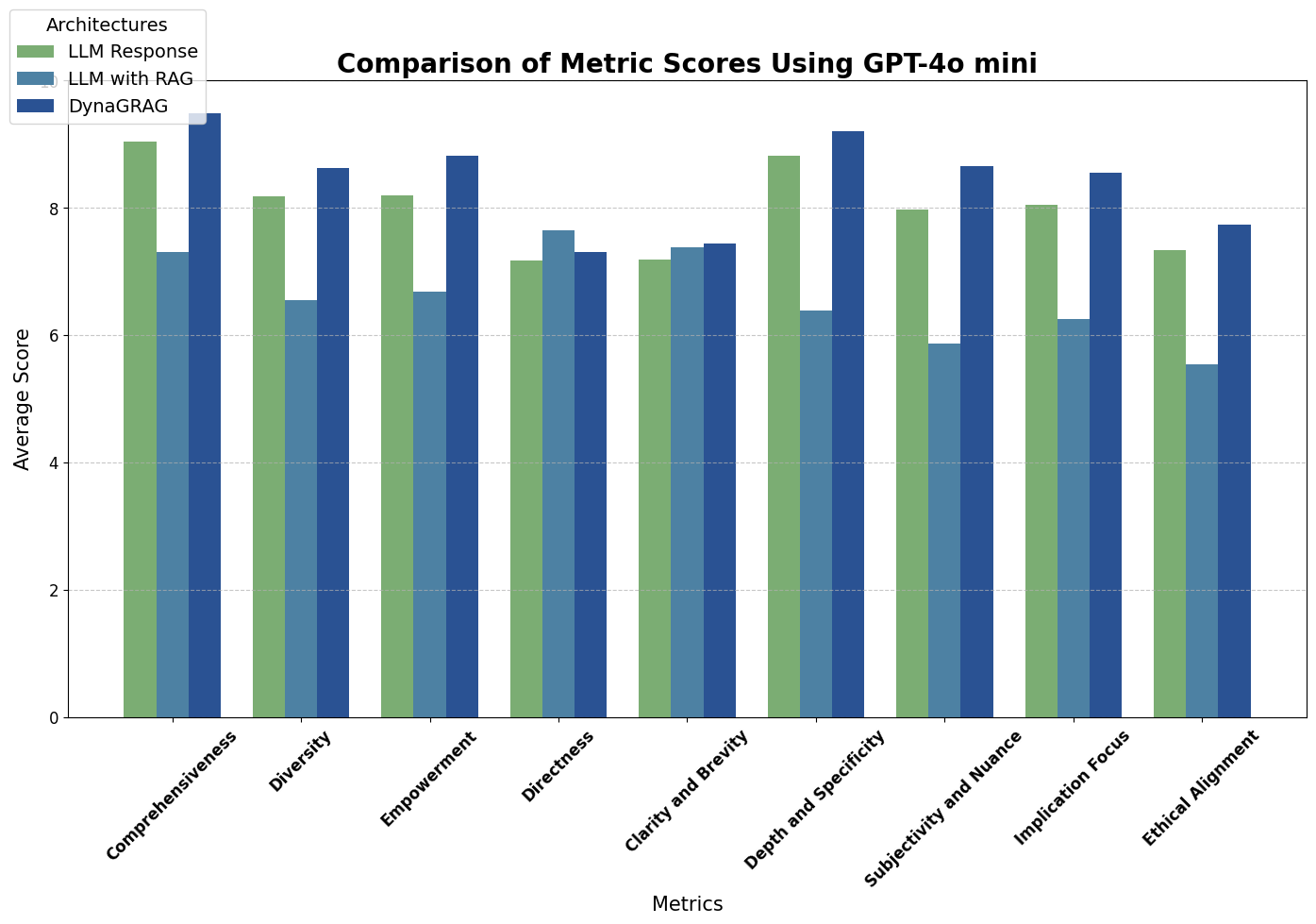}
        \caption{Bar plot of architecture results using GPT-4o mini.}
        \label{fig:openai_results}
    \end{minipage}
\end{figure}

\section{Discussion}

The DynaGRAG framework demonstrated significant advancements over state-of-the-art LLM pipelines by dynamically adapting to queries and uncovering complex relationships and contextual nuances that other methods, such as Vanilla LLMs or naïve retrieval-augmented pipelines, often struggled to capture—all without requiring additional LLM fine-tuning. While evaluation metrics like coherence and relevance provided a robust foundation for assessing responses, they tended to focus on whether a response satisfied specific criteria, rather than capturing the depth and quality of how well those criteria were met. This reflects a broader challenge in evaluating nuanced reasoning, as LLMs themselves can struggle to fully measure qualities like comprehensiveness or contextual alignment. Despite these evaluation constraints, DynaGRAG effectively leveraged its graph-based architecture to produce contextually rich and thoughtfully constructed outputs, demonstrating its ability to surpass high baseline scores and push the boundaries of what these systems could achieve.


The depth of DynaGRAG’s advancements becomes particularly evident when the responses are examined qualitatively. While the Vanilla LLM produced outputs that were coherent and satisfactory, DynaGRAG consistently went beyond the surface, delivering responses enriched with nuance, drawn from diverse contexts, and infused with deeper, more meaningful connections. As highlighted in the comparison tables, DynaGRAG’s outputs demonstrate a level of reasoning and contextual depth that is difficult to fully capture through metrics alone, further underscoring the innovative value of this architecture.


\begin{table}[H]
\centering
\renewcommand{\arraystretch}{1.2} 
\setlength{\tabcolsep}{5pt} 
\small 
\begin{tabular}{|l|p{13cm}|} 
\hline
\textbf{Architecture} & \textbf{Query: Based on the provided texts, what's the most exciting innovation in AI currently?} \\ \hline
\textbf{Vanilla LLM}\footnotemark[1] & According to Sholto Douglas, the most exciting innovation in AI is the ability to have long context lengths. He says it allows the model to ingest a massive amount of information, instantly solving the onboarding problem. \\ \hline
\textbf{Naïve RAG} & Dwarkesh Patel raises concerns about the lack of a null hypothesis when studying AI models and intelligence. He argues that this lack of a null hypothesis increases uncertainty because it makes it difficult to establish a baseline for comparison and to determine whether observed improvements are due to actual progress or simply random fluctuations. Without a null hypothesis, it becomes challenging to distinguish between meaningful advancements and mere chance occurrences. He believes that establishing a null hypothesis is crucial for more rigorous analysis and a better understanding of AI progress. \\ \hline
\textbf{DynaGRAG} & Determining the single "most exciting" innovation in AI is subjective and depends on individual priorities, but several areas are generating significant excitement within the field.  While advancements in Large Language Models (LLMs) like GPT and Gemini are undeniably impressive, showcasing human-level capabilities in certain tasks and fueling rapid development, the most compelling contender for the title of "most exciting" is arguably \textbf{mechanistic interpretability}.

The intense focus on mechanistic interpretability stems from the critical need to understand \textbf{how} these powerful LLMs function.  This is not merely a scientific challenge; it is crucial for ensuring the safe and responsible development of AI.  The high prominence given to model interpretability in the provided texts (with scores consistently above 8.0), along with the highlighted contributions of companies like Anthropic in openly publishing their research, emphasizes its importance.  Mechanistic interpretability directly addresses the "black box" problem, aiming to unveil the internal decision-making processes of LLMs, which is essential for mitigating risks associated with deception and unintended consequences.  This pursuit is intrinsically linked to other exciting areas, such as identifying "deception circuits" and developing automated interpretability tools.

While LLMs themselves represent a revolutionary leap in AI capabilities and their advancements continue to generate excitement, the advancements in techniques like \textbf{Reinforcement Learning from Human Feedback (RLHF)}, used to align LLMs with human values, also warrant mention.  Furthermore, the development of \textbf{multimodal AI}, capable of processing various data types, shows great promise and is actively researched.  The expansion of \textbf{context windows} in LLMs is another area of significant focus, promising further improvements in their abilities.

In summary, while the impressive capabilities of LLMs are undeniable, the race to understand their inner workings through mechanistic interpretability is arguably the most exciting current development. This pursuit is driven not only by scientific curiosity but also by the imperative to develop safe, reliable, and trustworthy AI systems as their power continues to grow.  The interconnectedness of these advancements – LLMs, RLHF, multimodal AI, and the crucial need for interpretability – highlights the dynamic and rapidly evolving nature of the AI field. \\ \hline
\end{tabular}
\caption{Responses provided by various architectures using Gemini 1.5 Flash. 
\textsuperscript{1}Due to the smaller dataset used for this test, the entire podcast transcript was included directly in the prompt.}
\label{tab:architecture_response}
\end{table}

\begin{table}[H]
\centering
\renewcommand{\arraystretch}{1.2} 
\setlength{\tabcolsep}{5pt} 
\small 
\begin{tabular}{|l|p{13cm}|} 
\hline
\textbf{Architecture} & \textbf{Query: What are the key unanswered questions regarding the nature of intelligence (both biological and artificial) that hinder our ability to accurately predict the future of AI development?} \\ \hline
\textbf{Vanilla LLM} & 
The nature of intelligence, both biological and artificial, remains shrouded in mystery, hindering our ability to accurately predict the future of AI development. Here are some key unanswered questions:
\textbf{1. What is Consciousness?} Biological: How does the human brain generate subjective experience? What is the relationship between neural activity and consciousness? Artificial: Can artificial intelligence ever truly be conscious? If so, how would we know? \textbf{2. How Does the Brain Learn and Generalize?} Biological: How does the human brain efficiently learn from limited data, generalize to new situations, and exhibit common sense reasoning? Artificial: Current AI models often struggle with generalization and require massive datasets. Can we develop AI systems that learn more like humans, with fewer data points and better adaptability?... \textbf{3. What are the Limits of Computation?}... \textbf{4. How Can We Ensure AI Safety and Control?}... \textbf{5. What is the Role of Embodiment and Interaction in Intelligence?}... These are just a few of the many unanswered questions that continue to challenge our understanding of intelligence and guide the future of AI development. Addressing these questions will require interdisciplinary research across fields such as neuroscience, cognitive science, computer science, and philosophy. \\ \hline
\textbf{Naïve RAG} & Based on the podcast transcript, several key unanswered questions hinder our ability to accurately predict AI development: \textbf{Regarding Biological Intelligence:} \textbf{The nature of reasoning and memory in the brain:}  While analogies are drawn between transformer models and the brain (e.g., residual streams and cerebellar circuits), the exact mechanisms of reasoning and how memories are accessed and integrated remain poorly understood.  The podcast highlights the debate on whether intelligence is primarily pattern matching or involves more complex deductive processes.  The relationship between memory, imagination, and reconstructive memory processes is also unclear. \textbf{The role of the cerebellum:}... \textbf{Regarding Artificial Intelligence:} \textbf{The true nature of in-context learning:}... \textbf{The potential for emergence:}... \\ \hline
\textbf{DynaGRAG} & Several key unanswered questions about the nature of intelligence, both biological and artificial, significantly hinder our ability to accurately predict the future of AI development. These uncertainties impact not only the technical trajectory of AI but also its economic viability and societal implications. One crucial area of uncertainty revolves around the \textbf{fundamental nature of intelligence itself}.  We lack a complete understanding of how consciousness arises in biological systems, and whether or not it is even a necessary component of intelligence.  This gap in knowledge makes it difficult to predict whether or when artificial systems might achieve consciousness, and what the implications of such a development would be.  Furthermore, we don't fully grasp the relationship between specialized functions and general intelligence, both in biological brains and in artificial systems.  The human brain's ability to integrate specialized regions into a unified, general intelligence remains a mystery, mirroring the challenges faced in creating truly generalizable AI.  Relatedly, the debate around whether all aspects of intelligence are reducible to computation, or whether non-computable elements exist, remains unresolved. This fundamental question bears directly on the potential for AI to ever truly replicate or exceed human-level intelligence. Another critical set of unanswered questions concerns the \textbf{scaling laws of intelligence}.  While empirical observations exist regarding the relationship between model size and performance in AI, the extrapolation of these scaling laws to larger models remains highly uncertain. We don't know if there are diminishing returns, thresholds, or critical points beyond which further scaling yields no significant improvement in intelligence.  Similarly, we lack a comprehensive understanding of how intelligence evolves biologically, what selection pressures favored the development of complex cognition, and how this understanding might inform the future development of AI through biologically-inspired approaches.The challenges extend to the \textbf{alignment problem}: how do we ensure that increasingly sophisticated AI systems remain aligned with human values? This question is profoundly linked to our understanding of intelligence itself.  If we don't fully understand how intelligence works, aligning advanced AI systems with human intentions becomes exponentially more difficult.  Furthermore, the pervasive issue of bias in AI systems, reflecting the biases of their creators and training data, underscores the need for a deeper understanding of how values shape AI development and its potential societal impact. Finally, the \textbf{definition of intelligence} itself is a source of uncertainty.  Is it purely efficient problem-solving across various modalities, or does it encompass subjective experience, consciousness, or self-awareness?  This ambiguity impacts predictions about future AI capabilities and fuels the ongoing debate around whether simply scaling Large Language Models (LLMs) will lead to Artificial General Intelligence (AGI), or if other, qualitatively different factors are at play. In summary, the inability to accurately predict the future of AI stems from a confluence of fundamental unknowns.  These include a lack of understanding regarding the very nature of intelligence (biological and artificial), the scaling laws governing its development, the challenges of aligning increasingly sophisticated AI systems with human values, and the persistent issue of bias. Addressing these questions is crucial not only for advancing AI technology but also for mitigating potential risks and harnessing its transformative power responsibly. \\ \hline
\end{tabular}
\caption{Responses provided by various architectures using Gemini 1.5 Flash}
\label{tab:architecture_response}
\end{table}

The performance of the architectures across the provided examples illustrates the nuanced advancements achieved by DynaGRAG in generating contextually rich and reasoning-intensive responses. While the Vanilla LLM and Naïve RAG pipelines produced responses that were coherent and relevant, they fell short in capturing the depth, diversity, and interconnectedness that DynaGRAG uniquely uncovered. For instance, when tasked with identifying the most exciting innovation in AI, the Vanilla LLM focused narrowly on context lengths, and Naïve RAG fixated on a single critical perspective about null hypotheses. In contrast, DynaGRAG synthesized a layered, multifaceted response, indicating how "the most exciting innovation" is highly subjective and a result of an individual's priorities. It also highlighted mechanistic interpretability, reinforcement learning from human feedback, and multimodal AI as promising avenues along with the overarching importance of interpretibility in shaping perceptions of technical advancements.

Similarly, in addressing key unanswered questions about intelligence, DynaGRAG far exceeded the scope of its counterparts by weaving together themes like the scaling laws of intelligence, the alignment problem, and the computational boundaries of intelligence. Unlike the Vanilla LLM’s more generalized exploration of key topics or the Naïve RAG’s surface-level framing, DynaGRAG integrated diverse sources to provide an expansive and deeply analytical response. It contextualized the technical and philosophical challenges of intelligence while maintaining clarity and coherence, demonstrating its ability to reason across complex, multifaceted queries.

What sets DynaGRAG apart is its capacity to illuminate connections that might otherwise remain obscured, delivering insights that feel both profound and purpose-driven. Its responses reflect a model capable of understanding not just the query but the broader narrative it seeks to address. This depth of reasoning aligns with one of the most critical themes in AI research: explainability. By leveraging graph-based retrieval and advanced prompting techniques, DynaGRAG not only unlocks a richer tapestry of knowledge but also offers a transparent lens into the context and sources used to derive its responses. 

Drawing from its ability to integrate diverse perspectives and contextual nuances through advanced reasoning, the result is a natural emergence of ethical and contextually aligned outputs, achieved without explicit mandates, demonstrating that robust reasoning capabilities can inherently foster alignment. For example, in the query about the most exciting AI innovations, DynaGRAG discussed several technical advancements like LLMs, RLHF, and multimodal AI and then tied these innovations to their broader implications for safety, transparency, and trustworthiness. Its response highlighted how the pursuit of mechanistic interpretability is driven not merely by scientific curiosity but by the imperative to develop AI systems that are safe and reliable as their capabilities continue to develop. This perspective, reflecting the interconnectedness of advancements in AI, exemplifies the transformative potential of AI systems to synthesize nuanced insights and deliver responses that inspire more thoughtful exploration of critical questions.

\subsection{Limitations and Future Directions}

This research sought to compare DynaGRAG with Microsoft’s Graph RAG to understand its strengths relative to a well-established alternative. While DynaGRAG employs a dynamic and adaptive framework distinct from Microsoft's Graph RAG, such a comparison would have provided valuable methodological insights. However, evaluating at a scale of 500k tokens presents significant challenges. Open-source implementations of Microsoft's implementation were difficult to adapt as platforms like Groq imposed strict token limits and Ollama lacked sufficient processing speed. Despite exploring additional options such as Gemini 2.0 Flash Experimental and the MultiHop-RAG benchmark, computational limitations remained a bottleneck.


As a solo developer, these constraints naturally influenced the scope of my evaluation. Future comparisons could benefit from narrowing the dataset to 100k tokens, allowing for more manageable testing of other architectures, including Microsoft's Graph RAG and open-source LLMs. Additionally, refining the approach to create hierarchical subgraphs could further enhance scalability and contextual understanding, enabling more efficient representation of complex relationships. Even so, DynaGRAG demonstrated significant advancements in reasoning and contextual depth over advanced models like Gemini 1.5 Flash while processing queries in approximately one minute and without any additional LLM finetuning. This efficiency highlights its practical viability and cost-effectiveness for real-world applications.


The experiments also underscored the critical role of graph density in enabling robust reasoning and context synthesis. Dense, well-connected graphs excel at capturing intricate relationships and synthesizing nuanced insights, particularly in smaller datasets where interconnections are more pronounced and noise is minimal. However, scaling to larger datasets introduces challenges, as sparse graphs dilute these connections and hinder information synthesis. Addressing this disparity through techniques to preserve graph density at scale is a critical area for future work. Such advancements could unlock the ability to maintain nuanced reasoning and context synthesis across diverse data regimes, further enhancing the scalability and adaptability of the framework.

\section{Conclusion}

This work introduces DynaGRAG as a meaningful step forward in advancing graph-based reasoning and language generation. By combining LLMs with graph learning techniques, DynaGRAG achieves an adaptive and query-aware capability, retrieving and synthesizing subgraphs to uncover complex relationships. The framework consistently outperformed baseline methods, demonstrating its strength in producing nuanced, contextually rich responses that address reasoning-intensive challenges. The ability to dynamically adapt to diverse queries without requiring fine-tuning emphasizes its scalability and practicality for real-world applications.

The broader implications of this research extend beyond technical performance, positioning DynaGRAG as a tool for fostering deeper understanding. As Yann LeCun has noted, LLMs are not yet capable of human-like reasoning~\cite{Hart:2024}, but intelligence itself remains an elusive concept without a complete definition. This ambiguity was evident in one of the tested queries, which explored the future of intelligence. DynaGRAG’s response to this query highlighted a profound duality: while we can reason about what we know, there remain aspects of intelligence we have yet to see or understand. By connecting themes like mechanistic interpretability, scaling laws, and alignment challenges, DynaGRAG demonstrated how reasoning systems can navigate these uncharted territories, pushing the boundaries of what AI can achieve in reasoning and understanding.

DynaGRAG moves beyond generating outputs by providing transparency into the context and relationships underpinning its responses, addressing the growing demand for explainability in AI systems. This capability fosters trust and alignment, enabling applications in areas like knowledge discovery, ethical decision-making, and solving complex problems where clarity is essential. By synthesizing diverse contexts, DynaGRAG uncovers hidden connections and exposes gaps in understanding, encouraging exploration of what remains unknown. Its emphasis on reasoning as a core aspect of intelligence offers not just answers but new perspectives, redefining how AI systems engage with knowledge. This work highlights the transformative role of advanced reasoning frameworks in expanding the boundaries of our understanding and shaping the future of intelligent systems.

\newpage
\setcounter{footnote}{1}
\bibliography{mybib} 

\end{document}